\documentclass[11pt]{article}
\usepackage{acl2016}
\usepackage{times}
\usepackage{url}
\usepackage{latexsym}
\usepackage{times}
\usepackage{latexsym}
\usepackage{amsmath}
\usepackage{multirow}
\usepackage{url}
\usepackage{graphicx}
\usepackage{caption}
\usepackage{subcaption}
\usepackage{tabularx, booktabs}
\makeatletter
\newcommand{\@BIBLABEL}{\@emptybiblabel}
\newcommand{\@emptybiblabel}[1]{}
\makeatother

\aclfinalcopy % Uncomment this line for the final submission

\title{Addressing Limited Data for Textual Entailment Across Domains}

\author{Chaitanya Shivade\Thanks{This work was conducted during an internship at IBM} \ \ \  Preethi Raghavan$^\dag$ \and Siddharth Patwardhan$^\dag$
  \vspace{10pt}
  \\
  $^\ast$Department of Computer Science and Engineering,\\
  The Ohio State University,\\
  Columbus, OH 43210\\
  {\tt shivade@cse.ohio-state.edu}
  \vspace{10pt}
  \\
  $^\dag$IBM T. J. Watson Research Center,\\
  1101 Kitchawan Road,\\
  Yorktown Heights, NY 10598\\
  {\tt \{praghav,siddharth\}@us.ibm.com}\\}

\date{}

\begin{document}

\maketitle

\begin{abstract}
We seek to address the lack of labeled data (and high cost of annotation) for textual entailment in some domains. To that end, we first create (for experimental purposes) an entailment dataset for the clinical domain, and a highly competitive supervised entailment system, {\sc Ent}, that is effective (out of the box) on two domains. We then explore self-training and active learning strategies to address the lack of labeled data. With self-training, we successfully exploit unlabeled data to improve over {\sc Ent} by 15\% F-score on the newswire domain, and 13\% F-score on clinical data. On the other hand, our active learning experiments demonstrate that we can match (and even beat) {\sc Ent} using only 6.6\% of the training data in the clinical domain, and only 5.8\% of the training data in the newswire domain.

%We describe an entailment system that incorporates characteristics of successful systems from the RTE challenges, along with additional features, and performs competitively in both newswire and clinical domains. Specifically, we address the problem of limited labaled data, in domains other than newswire, using self-training and active learning. We observe significant improvements with these approaches on datasets across both domains.

%We describe a highly configurable textual entailment system that can be easily adapted to new domains using domainspecific resources. 
%Specifically, we apply our system to entailment in the clinical domain,and we demonstrate superior performance.
%We investigate the problem of textual entailment search in the clinical domain where annotations are expensive, and document characteristics are substantially different from newswire text. 
%We present results for these approaches on a textual entailment dataset developed with clinical text, as well as the RTE-5 dataset, and compare them against multiple supervised baselines in both domains. We find that not only does the use of self-training result in significant improvements over a supervised baseline in the clinical domain, but it also achieves the best result among all systems that had participated in the RTE-5 search task. We further demonstrate that active learning reduces the number of required annotations by more than 90\% in both the clinical and RTE-5 datasets.
\end{abstract}

\section{Introduction}
\label{sec:intro}

Textual entailment is the task of automatically determining whether a natural language \textit{hypothesis} can be inferred from a given piece of natural language \textit{text}. The RTE challenges \cite{RTE5,Bentivogli2011} have spurred considerable research in textual entailment over newswire data. This, along with the availability of large-scale datasets labeled with entailment information \cite{bowman2015}, has resulted in a variety of approaches for textual {\em entailment recognition}. 

A variation of this task, dubbed textual \textit{entailment search}, has been the focus of RTE-5 and subsequent challenges, where the goal is to find all sentences in a corpus that entail a given hypothesis. The mindshare created by those challenges and the availability of the datasets has spurred many creative solutions to this problem. However, the evaluations have been restricted primarily to these datasets, which are in the newswire domain.
%Participating systems have incorporated alignment and matching based approaches using various lexical, syntactic and semantic similarity measures to determine entailment between hypotheses and text pairs. These approaches have met with moderate success for the entailment search task and have been used in a multitude of applications including multi-document summarization, question answering, machine translation and parser evaluation  \cite{dagan2013}.
Thus, much of the existing state-of-the-art research has focused on solutions that are effective in this domain.
%research and development of systems for addressing the entailment problem in the newswire domain.

It is easy to see though, that entailment search has potential applications in other domains too. For instance, in the clinical domain we imagine entailment search can be applied for clinical trial matching as one example.
%a patient's electronic medical records contain a large number of unstructured clinical notes. One such important application is in clinical trial matching, where 
Inclusion criteria for a clinical trial (for e.g., {\em patient is a smoker}) become the hypotheses, and the patient's electronic health records are the text for entailment search. Clearly, an effective textual entailment search system could possibly one day fully automate clinical trial matching.
%to be matched against, to determine eligibility for the trial.
%For instance, the validity of an inclusion criteria, \textit{Patient is a current smoker}, may be determined by matching it against the text in the patient's record where

%Hypothesis could be expressed in many different ways in the patient record \cite{Meystre2008} making it a major challenge to automate clinical trial matching.
Developing an entailment system that works well in the clinical domain and, thus, automates this matching process, requires lots of labeled data, which is extremely scant in the clinical domain. Generating such a dataset is tedious and costly, primarily because it requires medical domain expertise. Moreover, there are always privacy concerns in releasing such a dataset to the community. Taking this into consideration, we investigate the problem of textual entailment in a low-resource setting.

We begin by creating a dataset in the clinical domain, and a supervised entailment system that is competitive on multiple domains -- newswire as well as clinical. We then present our work on self-training and active learning to address the lack of a large-scale labeled dataset. Our self-training system results in significant gains in performance on clinical (+13\% F-score) and on newswire (+15\% F-score) data. Further, we show that active learning with uncertainty sampling reduces the number of required annotations for the entailment search task by more than 90\% in both domains.
%These results are promising since they generalize across both clinical and RTE data.

\section{Related work}
\label{sec:relatedwork}

Recognizing Textual Entailment (RTE) shared tasks \cite{dagan2013} conducted annually from 2006 up until 2011 have been the primary drivers of textual entailment research in recent years. Initially the task was defined as that of {\em entailment recognition}.
%Because of the initial focus of RTE on {\em entailment recognition}, most studies -- including recent work of \newcite{bowman2015} -- have investigated this type of entailment.
%However, the  the search task is comprised of issues such as interpreting sentences in their discourse context, understanding implicit and explicit references to entities, dates, time, etc.  making it harder than the recognition task \cite{RTE5}.
%Datasets were created for the first five RTE challenges by collecting text-hypothesis pairs for different NLP application scenarios. The challenge organizers ensured that the datasets had a class balance. Therefore, the distribution of entailment pairs in these datasets did not represent a \textit{natural} distribution of entailment cases in an actual application setting, where non-entailment pairs significantly outnumber entailment pairs \cite{dagan2013}. 
RTE-5 \cite{RTE5} then introduced the task of {\em entailment search} as a pilot.
%In this type of entailment, the goal of a system is to automatically find, within a set of documents on a topic, all sentences entailing a given hypothesis.
Subsequently, RTE-6  \cite{Bentivogli2010} and RTE-7 \cite{Bentivogli2011} featured entailment search as the primary task, but constrained the search space to only those candidate sentences that were first retrieved by Lucene, an open source search engine\footnote{\url{http://lucene.apache.org}}. Based on the 80\% recall from Lucene in RTE-5, the organizers of RTE-6 and RTE-7 deemed this filter to be an appropriate compromise between the size of the search space and the cost and complexity of the human annotation task.
%with a small difference: given a set of documents about a topic, a hypothesis H, and  a set of candidate entailing sentences (retrieved by an open source search engine\footnote{\url{http://lucene.apache.org/}} from these documents), the goal was to identify all sentences that entail H among the candidate sentences. The organizers found that Lucene search was able to achieve a recall of 80\% on the RTE-5 Pilot development set. Thus, in RTE-6 and RTE-7 the search task is only over the Lucene retrieved candidates \cite{Bentivogli2010}. The organizers deemed this to be a good compromise, providing sufficient number of entailing sentences while making the number of gold standard annotations manageable to obtain.

Annotating data for these tasks has remained a challenge since they were defined in the RTE challenges. Successful approaches for entailment \cite{Mirkin2009,Jia2010,Tsuchida2011} have relied on annotated data to either train classifiers, or to develop rules for detecting entailing sentences. Operating under the assumption that more labeled data would improve system performance, some researchers have sought to augment their training data with automatically or semi-automatically obtained labeled pairs \cite{Burger2005,Hickl2006,Hickl2007,Zanzotto2010,Celikyilmaz2009}.

% Some systems have chosen to automatically augment their training data with new entailment pairs
%Although annotations for textual entailment are costly and time-consuming, there have been very few attempts to automatically discover entailment pairs.
\newcite{Burger2005} automatically create an entailment recognition corpus using the news headline and the first paragraph of a news article as near-paraphrases. Their approach has an estimated accuracy of 70\%  on a held out set of 500 pairs. The primary limitation of the approach is that it only generates positive training examples. \newcite{Hickl2006} improves upon this work by including negative examples selected using heuristic rules (e.g., sentences connected by \textit{although}, \textit{otherwise}, and \textit{but}). On RTE-2 their method achieves accuracy improvements of upto 10\%. However, \newcite{Hickl2007} achieves only a 1\% accuracy improvement on RTE-3 using the same method, suggesting that it is not always as beneficial.

Recent work by \newcite{bowman2015} describes a method for generating large scale annotated datasets, viz., the Stanford Natural Language Inference (SNLI) Corpus, for the problem of entailment recognition. They use Amazon Mechanical Turk to very inexpensively produce a large entailment annotated data set from image captions.

\newcite{Zanzotto2010} create an entailment corpus using Wikipedia data. They hand-annotate original Wikipedia entries, and their associated revisions for entailment recognition. Using a previously published system for RTE \cite{zanzotto2006}, they show that their expanded corpus does not result in improvement for RTE-1, RTE-2 or RTE-3.

Similarly, \newcite{Celikyilmaz2009} address the lack of labeled data by semi-automatically creating an entailment corpus, which they use within their question answering system. They reuse text-hypothesis pairs from RTE challenges in addition to manually annotated pairs from a newswire corpus (with pairs for annotation obtained through a Lucene search over the corpus).
%To obtain unlabeled text-hypothesis pairs, the authors used Lucene to query a large newswire corpus with news headlines. Instead of choosing the first sentence of the document as in previous studies, they used the first and twentieth sentence of the search as positive and negative entailment, respectively. The accuracy of their QA system on TREC04 increased with adding unlabeled data.

%Our research seeks to further explore this direction of expanding the existing training data set through semi-supervised means, and reducing annotation costs through active learning strategies, both of which have not been applied previously to this task.

%Work done for textual entailment has been evaluated on controlled newswire datasets. Our work investigates the problem with real-world data. 
%There has been no work exploring semi-supervised techniques such as self-training in textual entailment. To the best of our knowledge, this is also the first work exploring the use of active learning techniques for textual entailment.

Note that all of the above research on expanding the labeled data for entailment has focused on {\em entailment recognition}. Our focus in this paper is on improving {\em entailment search} by exploiting unlabeled data with self-training and active learning.

\section{Datasets}
\label{sec:dataset}

%The main contribution of our work aims to address the challenges of limited textual entailment data in new domains. In many domains, such as the biomedical domain, obtaining such data can be fairly expensive, because of the specialized human expertise (for e.g., qualified medical professionals) required in producing such data.
In this section, we describe the data sets from two domains, {\em newswire} and {\em clinical}, that we use in the development and evaluation of our work.

\subsection{Newswire Domain}
For the newswire domain, we use entailment search data from the {\sc Pascal} RTE-5, RTE-6 and RTE-7 challenges \cite{RTE5,Bentivogli2010,Bentivogli2011}. The dataset consists of a corpus of news documents, along with a set of hypotheses. The hypotheses come from a separate summarization task, where the summary sentences about a news story (given a topic) were manually created by human annotators. These summary sentences are used as hypotheses in the dataset. Entailment annotations are then provided for a subset of sentences from the document corpus, based on a Lucene filter for each hypothesis.

%{\bf The following discussion is incorrect -- RTE-5 thru' RTE-7 used an Update Summarization task for data creation}. These datasets were created by human annotators who used relevant text snippets suggested by a QA system as a candidate answer and turned the question into an affirmative sentence with the candidate answer ``plugged in'' to form the hypothesis.
%{\bf The dataset was created by  -- doing what? -- ... It consists of how much data? ...}

In this work, we use the RTE-5 development data to train our system ({\em Newswire-train}), RTE-5 test data for evaluation of our systems ({\em Newswire-test}), and we use the combined RTE-6 development and test data for our system development and parameter estimation ({\em Newswire-dev}). We use all of the development and test data from RTE-7, without the human annotation labels, as our unlabeled data ({\em Newswire-unlabeled}) for self-training and active learning experiments. A summary of the newswire data is shown in Table \ref{tab:datasummary}.
%{\bf -- Check all of this!! - Its correct}

\begin{table}
\centering
\begin{tabular}{lcc}
\hline
%                   & \textbf{Total} &                    \\
\textbf{Dataset}   & \textbf{Size}  & \textbf{Entailing} \\
\hline
Newswire-train     & 20,104 & 810 (4.0\%) \\
Newswire-dev       & 35,927 & 1,842 (5.1\%) \\
Newswire-test      & 17,280 & 800 (4.6\%) \\
Newswire-unlabeled & 43,485 &  -          \\
\hline
Clinical-train     & 7,026   & 293 (4.1\%) \\
Clinical-dev       & 8,092   & 324 (4.0\%) \\
Clinical-test      & 10,466  & 596 (5.6\%) \\
Clinical-unlabeled & 623,600 &  -          \\
\hline
\end{tabular}
\caption{Summary of datasets}
\label{tab:datasummary}
\end{table}

%We begin by first outlining our annotated entailment dataset creation process. This process is partially driven by a clinical QA application %\cite{Ferrucci2012}, 
%under development.
%The work described in this paper is part of a larger effort to improve QA performance along the lines of \newcite{Harabagiu2006}. They demonstrate the effectiveness of textual entailment within their QA application.
%They report a significant increase in accuracy after a textual entailment system is used to re-rank candidate answers retrieved by the QA system. Thus,

\subsection{Clinical Domain}
\begin{figure}
	\noindent\fbox
	{
		\begin{minipage}{0.94\linewidth}
			\texttt{**NAME[XX (YY) ZZ]  has no liver problems.}\\
			\texttt{PAST MEDICAL HISTORY}\\					
			\texttt{1. Htn}\\			
			\texttt{Well controlled}\\
			\texttt{2. Diabetes mellitus}\\			
			\texttt{On regular dose of insulin.}\\
			\\
			\texttt{FAMILY HISTORY:}\\
			\texttt{Father with T2DM age unknown}
		\end{minipage}
	}
\caption{Excerpt from a sample clinical note}
\label{fig:sampleNote}
\end{figure}

There are no public datasets available for textual entailment \textit{search} in the clinical domain. %Recent work by \newcite{bowman2015} described a method for generating large scale annotated datasets for the problem of entailment recognition. However, our problem is that of entailment search. Therefore, we created a textual entailment search dataset over patient health data i.e. clinical notes (Figure \ref{fig:sampleNote}). 
In creating this dataset, we imagine a real-world clinical situation where hypotheses are facts about a patient that a physician seeing the patient might want to learn (e.g., {\em The patient underwent a surgical procedure within the last three months.}). The unstructured notes in the patients electronic medical record (EMR) is the text against which a system would determine the entailment status of the given hypotheses.

%Given a corpus of text, creating an entailment search dataset involves two parts: {\it hypotheses generation}, and {\it entailment annotation}.
Observe that the aforementioned real-world clinical scenario is very closely related to a question answering problem, where instead of hypotheses a physician may pose natural language questions seeking information about the patient (e.g., {\em Has this patient undergone a surgical procedure within the past three months?}). Answers to such questions are words, phrases or passages from the patient's EMR. Since we have access to a patient-specific question answering dataset over EMRs\footnote{a publication describing the question-answering dataset is currently under review at another venue} (henceforth, referred to as the QA dataset), we use it here as our starting point in constructing the clinical domain textual entailment dataset. 
% * <sidd@patwardhans.net> 2016-06-07T20:00:03.647Z:
%
% ^.
%To create such an entailment dataset, we begin with a existing dataset (QA dataset) of patient-specific questions along with their answers annotated in the patients' EHRs.

Given a question answering dataset, how might one go about creating a dataset on textual entailment? We follow a methodology similar to that of RTE-1 through RTE-5 for entailment set derived from question answering data.
The text corpus in our entailment dataset is the set of de-identified patient records associated with
%the QA application. Our hypotheses originate from the questions of
the QA dataset.
%To generate hypothese from a given question, the human annotators first chose a relevant text snippet (T) that was suggested by a QA system as a candidate answer \cite{Dagan2006}. They then turned the question into an affirmative sentence with the candidate answer ``plugged in'' to form the hypothesis (H). For example, given the question, ``Who is Ariel Sharon?'' and the candidate answer ``Israel's Prime Minister, Ariel Sharon, visited Prague'' (T), the hypothesis H is formed by turning the question into the statement ``Ariel Sharon is Israel's Prime Minister'', producing a positive entailment pair (H,T).
To generate hypotheses, human annotators converted questions into multiple assertive sentences, which is somewhat similar to what was done in the first five RTE challenges (RTE-1 through RTE-5). 
%with one difference. Neither did the annotators use a QA system, nor did they refer the patient's EHR to get candidate answers.  Instead, in order to generate hypotheses, 
For a given question, the human annotators plugged in clinically-plausible answers to convert the question into a statement that may or may not be true about a given patient. Table \ref{tab:hypotheses} shows example hypotheses and their source questions. Note that this procedure for hypothesis generation diverges slightly from the RTE procedure, where answers from a question answering system were plugged into the questions to produce assertive sentences.

{
\renewcommand{\arraystretch}{1.3}
\begin{table*}
\centering
\begin{tabular}{p{6cm}p{9cm}}
\hline
\multicolumn{1}{c}{\textbf{Question}} & \multicolumn{1}{c}{\textbf{Hypotheses}} \\
\hline
When was the patient diagnosed with dermatomyositis? & {\tt \small The patient was diagnosed with dermatomyositis two years ago.}\\
Any creatinine elevation? & {\tt \small Creatinine is elevated.}\\
& {\tt \small Creatinine is normal.}\\
Why were xrays done on the forearm and hand? & {\tt \small Xrays were done on the forearm and hand for suspected fracture.}\\
\hline
\end{tabular}
\caption{Example question $\rightarrow$ hypotheses mappings}
\label{tab:hypotheses}
\end{table*}
}

%\subsection{Entailment annotations}
%\label{sec:dataset-entailment}
To generate entailment annotations, we paired a hypothesis with every sentence in a subset of clinical notes of the EHR, and asked human annotators to determine if the note sentence enabled them to conclude an entailment relationship with the hypothesis. For example, the text: \textit{``The appearance is felt to be classic for early MS.''} entails the hypothesis: \textit{``She has multiple sclerosis''}. While in the RTE procedure, a Lucene search was used as a filter to limit the number of hypothesis-sentence pairs that are annotated, in our clinical dataset we limit the number of annotations by pairing each hypothesis only with sentences from EMR notes containing an answer to the original question in the QA dataset. 
%Two annotators who are medical experts participated in creating the dataset. We observed high inter-annotator agreement (Cohen's kappa of 0.81) on a sample of 2883 text-hypothesis pairs.

The entailment annotations were generated by two medical students with the help of the annotations generated for QA. 11 medical students created our QA dataset of 5696 questions over 71 patient records, of which 1747 questions have corresponding answers. This was generated intermittently over a period of 11 months. Given the QA dataset, the time taken to generate entailment annotations includes conversion of questions to hypotheses, and annotating entailment. While conversion of questions to hypotheses took approx. 2 hours for 20 questions, generating about 3000 hypothesis and text pairs took approx. 16 hours. 
%Both questions and answers were generated by the medical students using a process that simulates a physician asking patient-specific questions in a clinical setting.
%(1) took approximately 2 annotator hours to generate about 20 questions per EMR and (2) took ..  in case of (3), identifying EMR notes containing an answer took the it took approximately 16 hours to pair about 3000 sentences with the generated hypotheses and identify entailment. 

%we also leveraged the existing answers to match notes up with hypotheses generated from the questions
%{\bf how long did it take? who were the annotators? make the point that this is an expensive process, since medical professionals and medical students are doing this annotation.}

At the end of this process, we had a total of 243 hypotheses annotated against sentences from 380 clinical notes, to generate 25,584 text-hypothesis pairs. We split this into train, development and test sets, summarized in Table \ref{tab:datasummary}. Although we have a fairly limited number of labeled text-hypothesis pairs, we do have a large number of patient health records (besides the ones in the annotated set). We generated unlabeled data in the clinical domain, by pairing the hypotheses from our training data with sentences from a set of randomly sampled subset of health records outside of the annotated data. %A summary of the clinical data is shown on the right side of Table \ref{tab:datasummary}.

%Unlabeled text-hypothesis pairs, for use in the self-training and active learning systems, were generated by {\bf -- how were these unlabeled pairs generated?}

Datasets for the textual entailment search task are highly skewed towards the non-entailment class. Note that our clinical data, while smaller in size than the newswire data, maintains a similar class imbalance. 

\section{Supervised Entailment System}
\label{sec:baselines}

%To begin exploring approaches for augmenting entailment systems with unlabeled data,
%for domains where human annotations are expensive to obtain,
We begin by defining, in this section, our supervised entailment system (called {\sc Ent}) that is used as the basis of our self-training and active learning experiments. Our system draws upon characteristics and features of systems that have previously been successful in the RTE challenges in the newswire domain. We further enhance this system with new features targeting the clinical domain. The purpose of this section is to demonstrate, through an experimental comparison with other entailment systems, that {\sc Ent} is competitive on {\em both} domains, and is a reasonable supervised system to use in our investigations into self-training and active learning.

\subsection{System Description}

%These systems were built for data from RTE challenges. Therefore, they lack domain-specific knowledge and do not address issues to described in Section \ref{sec:challenges}. Therefore, we developed an entailment search system using supervised learning for clinical data. 

%\subsection{Supervised learning system}
%Our system is trained binary classifier, using a set of sophisticated features inspired by previously successful entailment systems. We train a logistic regression classifier, with ridge estimator (we use the Weka implementation\cite{Hall2009}), powered by features that attempt to capture evidence of entailment across the sentence-hypothesis pair at many different levels -- lexical, syntactic, and semantic, etc.

Top systems \cite{Tsuchida2011,Mirkin2009} in the RTE challenges have used various types of passage matching approaches in combination with machine learning for entailment. We follow along these lines, and design a classifier-based entailment system. For every text-hypothesis pair in the dataset we extract a feature vector representative of that pair. Then, using the training data, we train a classifier to make entailment decisions on unseen examples. In our system, we employ a logistic regression with ridge estimator (the Weka implementation \cite{Hall2009}), powered by a variety of passage matching features described below.

{
\renewcommand{\arraystretch}{1.4}
\begin{table}
\centering
\small
\begin{tabular}{lp{0.66\linewidth}}
\hline
{\em Exact} & String match, ignore case\\
{\em Multi-word} & Overlapping terms in multi-word token\\
{\em Head} & String match head of multi-word token\\
{\em Wikipedia} & Wikipedia redirects and disamb. pages\\
{\em Morphology} & Derivational morphology, e.g. {\em archaeological $\rightarrow$ archaeology}\\
{\em Date+Time} & Match normalized dates and times\\
{\em Verb resource} & Match verbs using WordNet, Moby thesaurus, manual resources\\
{\em UMLS} & Medical concept match using UMLS\\
{\em Translation} & Affix-rule-based translation of medical terms to layman terms\\
\hline
\end{tabular}
\caption{{\sc Ent} term matchers}
\label{tab:supmatchers}
\end{table}
}

Underlying many of our passage match features is a more fine-grained notion of ``term match''. {\em Term matchers} are a set of algorithms that attempt to match tokens (including multi-word tokens, such as {\em New York} or {\em heart attack}) across a pair of passages. One of the simplest examples of these is {\em exact string matcher}. A token in one text passage that matches exactly, character-for-character, with a token in another text passage would be considered a term match by this simple term matcher. However, these term matchers could be more sophisticated and match pairs of terms that are synonyms, or paraphrases, or equivalent to one another according to other criteria. {\sc Ent} employs a series of term matchers listed in Table \ref{tab:supmatchers}. Each of these may also produce a confidence score for every match they find. Because we are working with clinical data, we added some medical domain term matchers as well -- using UMLS \cite{Bodenreider2004} and a rule-based ``translator'' of medical terms to layman terms\footnote{Rules for medical term translator were derived from {\tt http://www.globalrph.com/medterm.htm}}.

Listed below are all of our features used in the {\sc Ent}'s classifier. Most passage match features aggregate the output of the term matchers along various linguistic dimensions -- lexical, syntactic, semantic, and document/passage characteristics.

\noindent\underline{\bf Lexical:} This set includes a feature aggregating {\em exact string matches} across text-hypothesis, one aggregating {\em all term matchers}, a feature counting {\em skip-bigram matches} (using all matchers), a measure of {\em matched term coverage of text} (ratio of matched terms to unmatched terms). Additionally, we have some medical domain features, viz. {\em UMLS concept overlap}, and a measure of {\em UMLS-based similarity} \cite{Shivade2015,Pedersen2007} using the UMLS::Similarity tool \cite{McInnes2009}.
%, powered by features that attempt to capture evidence of entailment across the sentence-hypothesis pair at many different levels.
%-- lexical, syntactic, and semantic, etc. -- inspired by characteristics of previously successful entailment systems.

\noindent\underline{\bf Syntactic:} Following the lead of several approaches textual entailment \cite{wang2009,Mirkin2009,Kouylekov2010} we have a features measuring the {\em similarity of parse trees}. Our rule-based syntactic parser \cite{mccord89} produces dependency parses the text-hypothesis pair, whose nodes are aligned using all of the term matchers. The tree match feature is an aggregation of the aligned subgraphs in the tree (somewhat similar to a tree kernel \cite{moschitti06}). % Parse distance goes here... but looks like it only applies to focus+candidate passages.

\noindent\underline{\bf Semantic:} We apply open domain as well as medical entity and relation detectors \cite{wang11,wang12} to the texts, and post features measuring {\em overlap in detected entities} and {\em overlap in the detected relations} across the text-hypothesis pair. We also have a rule-based semantic frame detector for a ``medical finding'' frame (patient presenting with symptom or disease). We post a feature that aggregates matched {\em elements of detected frames}.

\noindent\underline{\bf Passage Characteristics}: Clinical notes typically have a structure and the content is often organized in sections (e.g. \textit{History of Illness} followed by \textit{Physical Examination} and ending with \textit{Assessment and Plan}). We identified the section in which each note sentence was located and used them as features in the classifier. Clinical notes are also classified into many different categories (e.g., {\em discharge summary}, {\em radiology report}, etc.), which we generate features from. We also generate several features capturing the ``readability'' of the text segments -- {\em parse failure}, {\em list detector}, {\em number of verbs}, {\em word capitalization}, {\em no punctuation} and {\em sentence size}. We also have {\em a measure of passage topic relevance} based on medical concepts in the pair of texts.

%\noindent{\bf UMLS Similarity Features}: UMLS::Similarity \cite{McInnes2009} is a freely available open source tool\footnote{\url{http://metacpan.org/pod/UMLS::Similarity}} that computes similarity between any two concepts from the UMLS. Given one or more ontologies, and one or more hierarchical relations of interest, the tool represents the ontologies as a graph, with concepts as nodes and relations as edges. We follow  \newcite{Shivade2015} and \newcite{Pedersen2007} who have shown that using Systematized Nomenclature of Medicine - Clinical Terms (SNOMED-CT) as the ontology with parent-child relationships between concepts is useful for similar tasks with clinical text in the past.

\subsection{System Performance}

\begin{table*}[ht]
\centering
\begin{tabular}{l|ccc|ccc}
\hline
& \multicolumn{3}{c|}{\textbf{Newswire}} & \multicolumn{3}{c}{\textbf{Clinical}} \\
\multicolumn{1}{c|}{\bf System} & \textbf{Precision} & \textbf{Recall} & \textbf{F-score} & \textbf{Precision} & \textbf{Recall} & \textbf{F-score} \\
\hline
Lucene      & 0.47       & 0.48       & {\bf 0.47}$^*$ & 0.16   & 0.22       & 0.19\\
{\sc Edits} & 0.22       & 0.57       & 0.32       & 0.23       & 0.21       & 0.20\\
{\sc Tie}   & 0.66       & 0.21       & 0.31       & 0.43       & 0.01       & 0.02\\
{\sc Ent}   & 0.77       & 0.26       & 0.39       & 0.42       & 0.15       & {\bf 0.23}$^*$\\
\hline
\end{tabular}
\caption{System performance on test data (* indicates statistical significance)}
\label{tab:system_results}
\end{table*}

%Many systems have been built by research teams around the world to address the RTE entailment search challenges (RTE-5 through RTE-7). Of these
%, and built a third supervised system (that we call {\sc Watson}) drawing upon the characteristics of some of best systems that participated in the RTE challenges.
%We compared our system with a number of baselines.  These are entailment systems that participated in previous RTE challenges and are freely available for download.
% \subsection{\sc Tie}
To compare effectiveness of {\sc Ent} on the entailment task, we chose two publicly available systems -- {\sc Edits} and {\sc Tie} -- for comparison. Both these system are available under the Excitement Open Platform (EOP), an initiative \cite{Magnini2014} to make tools for textual entailment freely available\footnote{\url{http://hltfbk.github.io/Excitement-Open-Platform/}} to the NLP community.
{\sc Edits} (Edit Distance Textual Entailment Suite) by \newcite{Kouylekov2010} is an open source textual entailment system that uses a set of rules and resources to perform ``edit'' operations on the text to convert it into the hypothesis. There are costs associated with the operations, and an overall cost is computed for the text-hypothesis pair, which determines the decision for that pair. This system has placed third (out of eight teams) in RTE-5, and seventh (out of thirteen teams) in RTE-7.
The Textual Inference Engine ({\sc Tie}) \cite{wang2009} is a maximum entropy based entailment system relying on predicate argument structure matching. While this system did not participate in the RTE challenges, it has been shown to be effective on the RTE datasets. In our experiments, we trained the {\sc Edits} system optimizing for F-score (the default optimization criterion is accuracy) and {\sc Tie} with its default settings. We also used a Lucene baseline similar to the one used in RTE-5, RTE-6 and RTE-7 entailment challenges.

We trained the systems on the training set of each domain and tested on the test set. The Lucene baseline considers the first $N$ sentences (where $N$ is 5, 10, 15 or 20) top-ranked by the search engine to be entailing the hypothesis. The configuration with the top 10 sentences performed the best, and is reported in the results. Note that this baseline is a strong one, and none of the systems participating in RTE-5 could beat it.
% \subsection{\sc Edits}
%\textbf{Lucene}: An open source search engine that was used as a baseline for the RTE-5 Pilot task. We used the exact same configuration as specified by the organizers \cite{RTE5}, but with a newer version of Lucene (v5.0) for our experiments.

Table \ref{tab:system_results} summarizes the system performance on newswire and clinical data. We observe that systems that did well on RTE datasets, were mediocre on the clinical dataset. We did not, however, put any effort into adaption of {\sc Tie} and {\sc Edits} to the clinical data. So the mediocre performance on clinical is understandable. It is interesting to see though that {\sc Ent} did well (comparatively) on both domains.
%However our supervised system had a robust performance, and outperformed others in both domains. Further, we conducted experiments using self-training and active learning on both the datasets. The following sections describe these results in detail.

%\subsection{Supervised baselines with RTE-5 data}
%\label{sec:rteexperiments}
We note that our problem setting is most similar to the RTE-5 entailment search task. 
Of the 20 runs across eight teams that participated in RTE-5, the median F-Score was 0.30 and the best system \cite{Mirkin2009} achieved an F-Score of 0.46.
%Table \ref{tab:rte5results} summarizes the results of our experiments on the test set.
{\sc Edits} and {\sc Tie} perform slightly above the median and {\sc Ent} (with 0.39 F-score) would have ranked third in the challenge.

The performance of all systems on the clinical data is noticeably low as compared to the newswire data. An obvious difference in the two domains is the training data size (see Table \ref{tab:datasummary}). However, obtaining annotations for textual entailment search is expensive, particularly in the clinical domain. The remaining sections present our investigations into self-training and active learning, to overcome the lack of training data.
%Semi-supervised and active learning techniques are attractive solutions in such situations. We discuss our experiments using these methods in the following sections.

\section{Self-Training}
\label{subsec:self-training}

Our goal is to exploit unlabeled data, with the hope of augmenting the limited annotated data in a given domain.
%available to train our machine learning model.
%To that end, we describe in this section, our self-training algorithm for learning from unlabeled data.
Self-training is a method that has been successfully used to address limited training data on many NLP tasks, such as parsing \cite{mcclosky2006effective}, information extraction \cite{huang2012bootstrapped,patwardhan2007ie}, word sense disambiguation \cite{mihalcea2004co}, etc. Self-training iteratively increases the size of the training set, by automatically assigning labels to unlabeled examples, using a model trained in a previous iteration of the self-training regime. % Figure \ref{fig:selftrain} shows a block schematic of this process.

%Self-training is a single-view weakly supervised algorithm that has been widely used in the NLP community \cite{sogaard2013}. In a standard supervised setup a model $M$ using the training set $L$, and tested it on the test set $T$. In self-training, an existing model labels unlabeled data. This newly labeled data is then treated as truth, and combined with the actual labeled data to train a new model.  

For our newswire and clinical datasets, using the set of unlabeled text-hypothesis pairs $U$, we ran the following training regime: A model was created using the training data $L_n$, and applied it to the unlabeled data $U$. From $U$, all such pairs that were classified by the model as entailing pairs with high confidence (above a threshold $\tau$) were added to the labeled training data $L_n$ to generate $L_{n+1}$. Non-entailing pairs were ignored. A new model is trained on data $L_{n+1}$, and the above process repeated iteratively, until a stopping criteria is reached (in our case, all pairs from $U$ are exhausted).

%confidence posterior probability for the entailment class (note: we ignore the non-entailment class) was greater than a chosen threshold $\tau$. These pairs were then added to the training set $L$, which resulted in a bootstrapped training set $L'$. A new model $M'$ was then trained on $L'$ and tested on $T$.  This was repeated iteratively until all pairs from U were exhausted.
%We observed significant gains using self-training with both the RTE data, and our clinical data. %The following subsections describe these experiments.

%\subsection{Self-Training on Clinical Domain}
%\label{sec:clinical-semisupervised}

%The performance (F-Score) obtained in a supervised setup on the development and the test sets is summarized in Table \ref{tab:clinical_results}. We consider the performance of our model in this setup as the baseline, and investigate if the use of self-training is beneficial.  Although we had a limited number of annotated clinical notes, we had a large number of unlabeled ones. We randomly sampled 10,000 clinical notes from our data, and paired each sentence with every hypotheses created for the training dataset. This gave us a total of 623,600 text-hypothesis pairs.

\begin{figure*}
	\begin{subfigure}{.5\textwidth}
	 	 \includegraphics[scale=0.25]{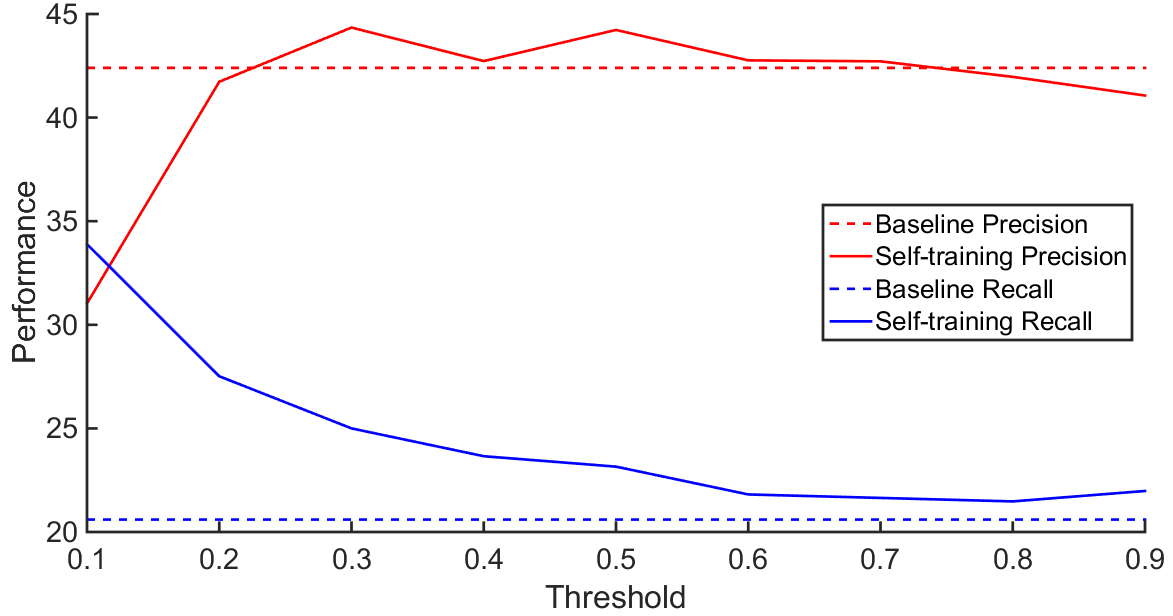}
	 	 \caption{Precision and recall for clinical data}
	  	\label{fig:clinical-ss-pr}
	\end{subfigure}
\hfill
	\begin{subfigure}{.5\textwidth}
		  \includegraphics[scale=0.25]{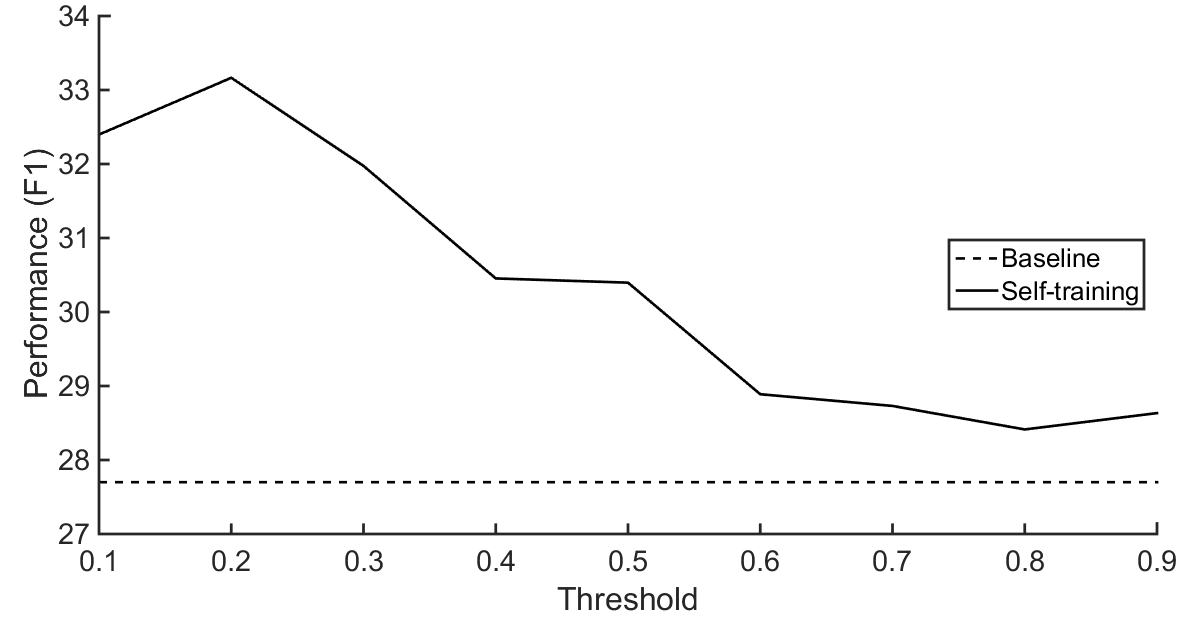}
		  \caption{F-Score for clinical data}
		  \label{fig:clinical-ss-f1}
	\end{subfigure}
    \begin{subfigure}{.5\textwidth}
	 	 \includegraphics[scale=0.25]		 {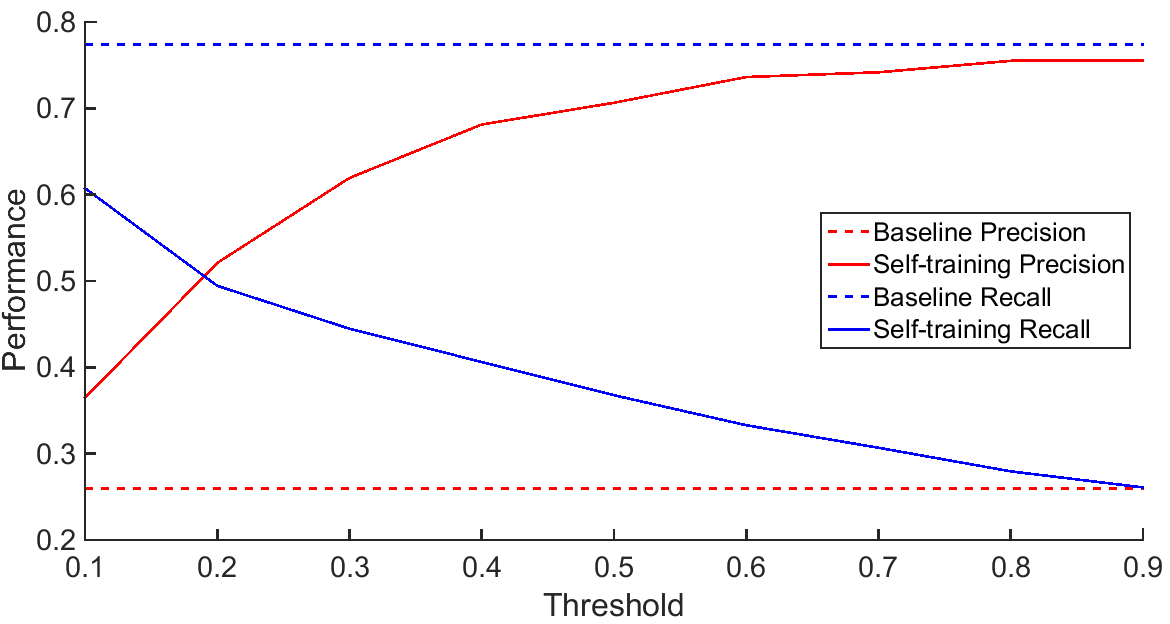}
	 	 \caption{Precision and recall for newswire data}
	  	\label{fig:rte-ss-pr}
	\end{subfigure}
\hfill
	\begin{subfigure}{.5\textwidth}
		  \includegraphics[scale=0.25]{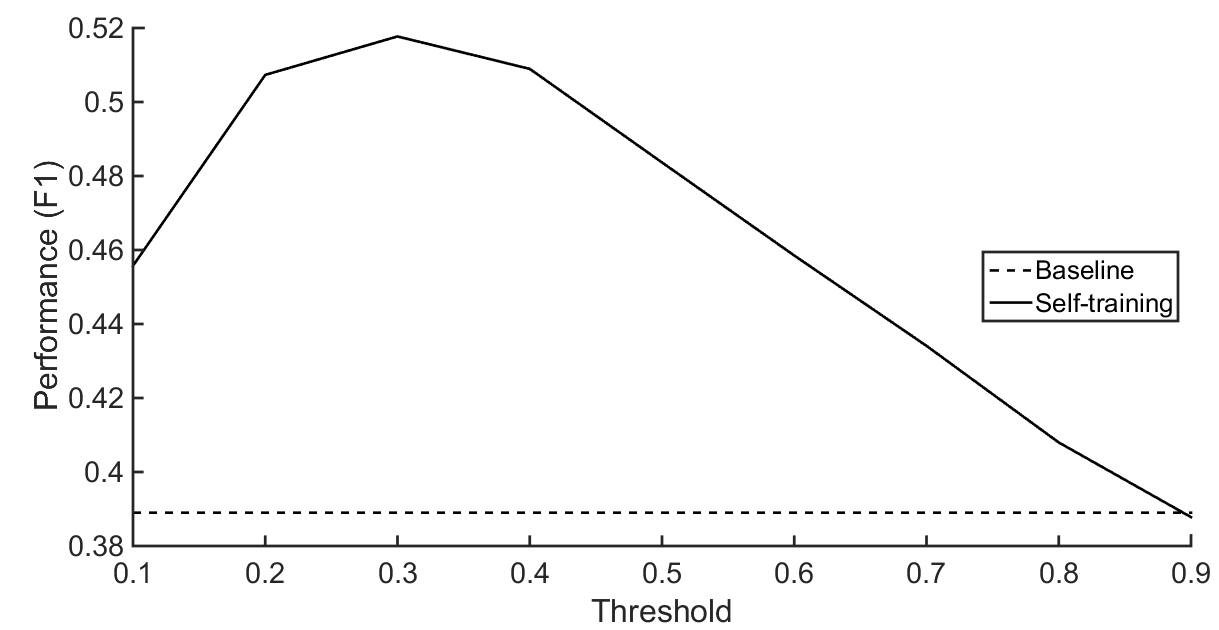}
		  \caption{F-Score for newswire data}
		  \label{fig:rte-ss-f1}
	\end{subfigure}
\caption{Self-training on development data}
\label{fig:semisupervised}
\end{figure*}

The threshold $\tau$ determines the confidence of our model for a text-hypothesis pair being classified to the entailment class. This threshold was tuned by varying it incrementally from 0.1 to 0.9 in steps of 0.1. The best $\tau$ was determined on the development set, and chosen for the self-training system. Figure \ref{fig:semisupervised} shows the effect of $\tau$ on the development data.
%Figure \ref{fig:clinical-ss-pr} shows changes in precision and recall on the development set as threshold $\tau$ changes. We observe that as $\tau$ decreases from 0.9 to 0.1, the precision decreases, while the recall increases. 

\begin{table*}[ht]
\centering
\begin{tabular}{l|ccc|ccc}
\hline
& \multicolumn{3}{c|}{\textbf{Newswire}} & \multicolumn{3}{c}{\textbf{Clinical}}\\
%\hline
\multicolumn{1}{c|}{\bf System} & \textbf{Precision} & \textbf{Recall} & \textbf{F-score} & \textbf{Precision} & \textbf{Recall} & \textbf{F-score} \\
\hline
{\sc Ent} & 0.77 & 0.26 & 0.39 & 0.42 & 0.15 & 0.23\\
{\sc Ent} + Self-Training & 0.62 & 0.48 & \textbf{0.54}$^*$ & 0.34 & 0.39 & \textbf{0.36}$^*$\\
\hline
\end{tabular}
\caption{Self-training results on test data (* indicates statistical significance)}
\label{tab:ss_results}
\end{table*}

As such, we see that the F-score of the self-trained model is always above that of the baseline {\sc Ent} system. The F-score increases upto a peak of 0.33 at threshold $\tau$ of 0.2 before dropping at higher thresholds.
%which is significantly greater than the performance obtained with the supervised baseline (F=27.8).
Using this tuned threshold on test set, the comparitive performance on the test set is outlined in Table \ref{tab:ss_results}.
We observe an F-score of 0.36, which is significantly greater than that of the vanilla {\sc Ent} system (0.23).

%The variation in precision and recall as the threshold changes is very intuitive.
The effect of the threshold on performance correlates with the number of instances added to the training set. When the threshold is low, there are more instances being added  (10,799 at threshold of 0.1) into the training set. Therefore, recall is likely to benefit, since the model is exposed to a larger variety of text-hypothesis pairs. However, the precision is low since noisy pairs are likely to be added. When the threshold is high, fewer instances are added (316 at threshold of 0.9). These are the ones that the model is most certain about, suggesting that these are likely to be less noisy. Therefore, the precision is comparatively high.
%One can observe in Table \ref{tab:ss_results} that gains obtained from self-training are due to recall.

%\subsection{Self-Training on Newswire Domain}
%\label{sec:newswire-semisupervised}

We also ran our self-training approach on the Newswire datasets.
%In our experiments, we consider RTE-5 \emph{train} as the training set and RTE-6 \emph{test} as our development set (43,845 text hypothesis pairs). We treat RTE-7 as our unlabeled set (43,845 text hypothesis pairs) by ignoring the labels. The threshold $\tau$ was tuned on the development set tested on the RTE-5 test set.
We observed similar variations in performance with newswire data as with the clinical data.  At threshold of 0.9, fewer instances (49) are added to the training set from the unlabeled data, while a large number of instances (2,861) are added at a lower threshold $\tau$ of 0.1. The best performance (F-score of 0.52) was obtained at threshold of 0.3, on the development set.

This threshold also resulted in the best performance (0.54) on the test set. Similar to the clinical domain, precision increased but recall decreased as the threshold increased. Again, it is evident from Table \ref{tab:ss_results} that gains obtained from self-training are due to recall. It should be noted that the self-trained system achieves an F-score of 0.54 -- substantially better than the best performing system of \newcite{Mirkin2009} (F-score, 0.46) in RTE-5.

\section{Active Learning}
\label{subsec:active-training}

%In this section, we describe our active learning approach to minimize the amount of labeled data required to train the machine learning models.

\begin{figure*}[t]
\centering
	\begin{subfigure}{0.49\textwidth}
		  \includegraphics[scale=0.25]{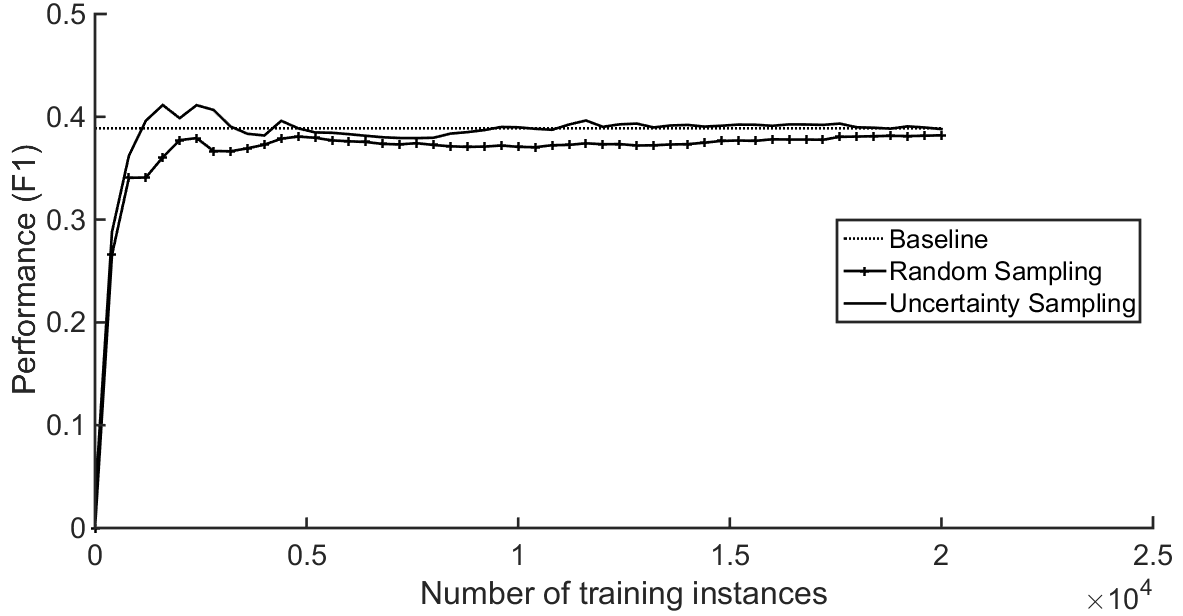}
		  \caption{Newswire Data}
		  \label{fig:active_rte}
	\end{subfigure}
\hfill
	\begin{subfigure}{.49\textwidth}
	 	 \includegraphics[scale=0.25]{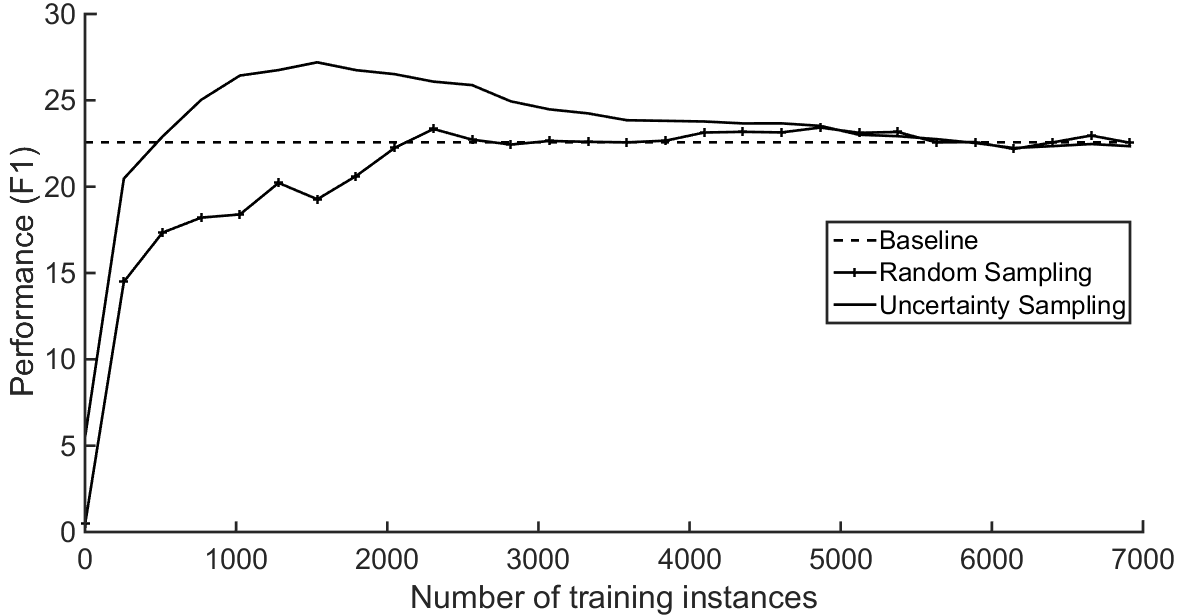}
	 	 \caption{Clinical Data}
	  	\label{fig:active_clinical}
	\end{subfigure}%

\caption{Learning curves for uncertainty sampling and random sampling on test data}
\label{fig:active}
\end{figure*}

Active learning is a popular training paradigm in machine learning \cite{settles2012} where a learning agent interacts with its environment in acquiring a training set, rather than passively receiving independent samples from an underlying distribution. This is especially pertinent in the clinical domain, where input from a medical professional should be sought only when really necessary, because of the high cost of such input. The purpose of exploring this paradigm is to achieve the best possible generalization performance at the lowest cost.
%, where cost is usually measured as a function of the number of labeled examples.
%Active learning has also been a topic of interest for a variety of problems in NLP \cite{olsson2009}.

Active learning is an iterative process, and typically works as follows: a model $M$ is trained using a minimal training dataset $L$. A query framework is used to identify an instance from an unlabeled set $U$ that, if added to $L$, will result in maximum expected benefit. Gold standard annotations are obtained for this instance and added to the original training set $L$ to generate a new training set $L'$. In the next iteration, a new model $M'$ is trained using $L'$ and used to identify the next most beneficial instance for the training set $L'$. This is repeated until a stopping criterion is met. This approach is often \textit{simulated} using a training dataset $L$ of reasonable size. The initial model $M$ is created using a subset $A$ of $L$. Further, instead of querying a large unlabeled set $U$, the remaining training data $(L-A)$ is treated as an unlabeled dataset and queried for the most beneficial addition. 

We carried out active learning in this setting using a querying framework known as \textit{uncertainty sampling} \cite{Lewis94}. Here, the model $M$ trained using $A$, queries the instances in ($L-A$) for instance(s) it is least certain for a prediction label. For probabilistic classifiers the most uncertain instance is the one where  posterior probability for a given class is nearest to 0.5. To estimate the effectiveness of this framework, it is always compared with a \textit{random sampling} framework, where random instances from the training data are incrementally added to the model.

%\subsection{Active Learning with Clinical Data}
%\label{sec:active_clinical}

%Our training data consists of 7,026 elements (refer to Table \ref{tab:clinicaldata}). The performance (F-Score) obtained using this data in a fully supervised setup on the development and test sets is summarized in Table \ref{tab:clinical_results}. We consider the performance of our model in this setup as the baseline, and investigate whether use of active learning is beneficial. 
Starting with a model trained using a single randomly chosen instance, we carried out active learning using uncertainty sampling, adding one instance at a time.
%Thus, we created 7,025 models until all training instances were added to the model.
After the addition of each instance, the model was retrained and tested on a held out set. To minimize the effect of randomization associated with the first instance, we repeated the experiment ten times and averaged the performance scores across the ten runs.  

Following previous work \cite{Settles2008,reichart2008} we evaluate active learning using learning curves on the test set. Figure \ref{fig:active} shows the learning curves for newswire and clinical data.

On clinical data, uncertainty sampling achieves a performance equal to the baseline {\sc Ent} with only 470 instances. With random sampling, over 2,200 instances are required. The active learner matches the performance of the {\sc Ent} with only 6.6\% of training data. Newswire shows a similar trend, with both sampling strategies outperforming {\sc Ent}, using less than half the training, and uncertainty sampling learning faster than random. While uncertainty sampling matches {\sc Ent} F-score with only 1,169 instances, random sampling requires 2,305. Here, the active learner matches {\sc Ent} performance using only 5.8\% of the training data.

%\subsection{Active Learning with Newswire Data}
%\label{sec:active_rte}

%In order to confirm the benefits of active learning, we repeated the experiments with RTE-5 Search Task data. The F-Score obtained by our model using this data in a fully supervised setup, is summarized in Table \ref{tab:rte5results}. This F-Score is considered as the baseline. We carried out both uncertainty sampling and random sampling using the RTE-5 test set as held out data. As with clinical data, the experiment was repeated ten times to and the results were averaged across the ten runs. These results are shown using learning curves in Figure \ref{fig:active_rte}. 

%\begin{table}
%\centering
%\begin{tabular}{cccc}
%\toprule
%\textbf{Data} & \textbf{Supervised} & \textbf{Active}	& \textbf{Annotation} \\
%		       &	\textbf{Baseline}	 &  \textbf{Learner}	& \textbf{Savings (\%)} \\
%\midrule
%Clinical		& 171	& 122	& 293	\\
%RTE			& 264	& 60		& 324	\\
%\bottomrule
%\end{tabular}
%\caption{Number of training instances required to reach optimal performance}
%\label{tab:active}
%\end{table}

%We found that active learning gave promising results on both the clinical and the RTE datasets. %The following sections describe these experiments in detail.

%\input{experiments}

%\input{semisupervised}

%\input{active}

\section{Effect of Class Distribution}
\label{sec:discussion}

After analyzing our experimental results, we considered that one possible explanation for the improvements over baseline {\sc Ent} could plausibly be because of changes in the class distribution.
%An important concern in training models for entailment search is the problem of class imbalance as it may have a significantly negative impact on the model.
From Table \ref{tab:datasummary}, we observe that the distribution of classes in both domains is highly skewed (only 4-5\% positive instances). Self-training and active learning dramatically change the class distribution in training. To assess the effect of class distribution changes on performance, we ran additional experiments, described here.

%\subsection{Class imbalance and sampling}
%\label{subsec:sampling}

We first investigated sub-sampling \cite{japkowicz2000learning} the training data to address class imbalance. This includes down-sampling the majority class or up-sampling the minority class until the classes are balanced. We found no significant gains over the vanilla {\sc Ent} baseline with both strategies. Specifically, down-sampling resulted in gains of only 0.002 and 0.001 F-score and up-sampling resulted in a drop of 0.011 and 0.013 F-score on clinical-dev and newswire-dev, respectively.

\begin{figure}[t]
 \includegraphics[scale=0.23]{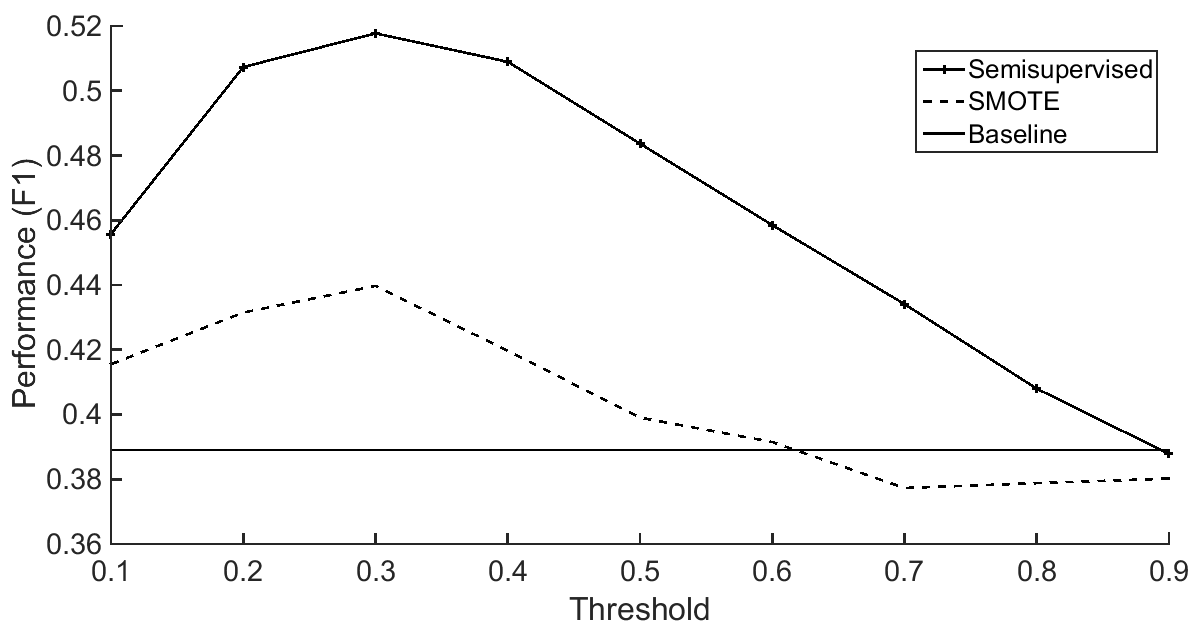}
\caption{Comparison of SMOTE and self-training (on newswire development set)}
\label{fig:smote}
\end{figure}

%Re-sampling the original dataset is a commonly used approach to address the problem of class imbalance \cite{japkowicz2000learning}. This involves down-sampling of the majority class or up-sampling of the minority class. We tried both, but did not get significant improvements over the supervised baseline. More specifically, downsampling the majority class until both classes were balanced resulted in gains of +0.002 and +0.001, and upsampling the minority class until balance resulted in a drop of -0.011 and -0.013 on the clinical and RTE devlopment sets respectively.
%\subsection{Comparison of synthetic oversampling and self-training}
%\label{subsec:smote}
Another approach to addressing class imbalance is to apply Synthetic Minority Oversampling Technique (SMOTE) \cite{chawla2002smote}. SMOTE creates instances of the minority class by taking a minority class sample and introducing synthetic examples between its $k$ nearest neighbors. Using SMOTE on newswire and clinical datasets resulted in improvements over baseline {\sc Ent} in both domains. The improvements using self-training, however, are significantly higher than SMOTE. Figure \ref{fig:smote} shows a comparison of SMOTE and self-training on newswire data, where equal number of instances are added to the training set by both techniques.
%In our experiments with self-training, we used the logistic regression model to extract positive examples from unlabeled data using a bootstrapping approach. In these experiments the model picks up examples from unlabeled data that are similar to the positive examples from its training set. Addition of positive examples also helps the model since the class balance is heavily skewed towards negative examples (see Section \ref{sec:dataset}). 
%Synthetic Minority Oversampling Technique (SMOTE) is an algorithm \cite{chawla2002smote} that addresses this problem by creating instances of the minority class in an imbalanced dataset. 
%These extra training samples are created by taking a minority class sample and  introducing synthetic examples along the line segments joining its $k$ nearest neighbors. 
%We used the default $k=5$ and compared SMOTE with our self-training model. 
%We found that while SMOTE results in improvement over the baseline supervised model, gains from self-training are significantly higher. Figure \ref{fig:smote} shows a comparison of SMOTE and self-training on RTE data, where equal number of instances are added to the training set by both the techniques.

\begin{figure}[t]
 \includegraphics[scale=0.28]{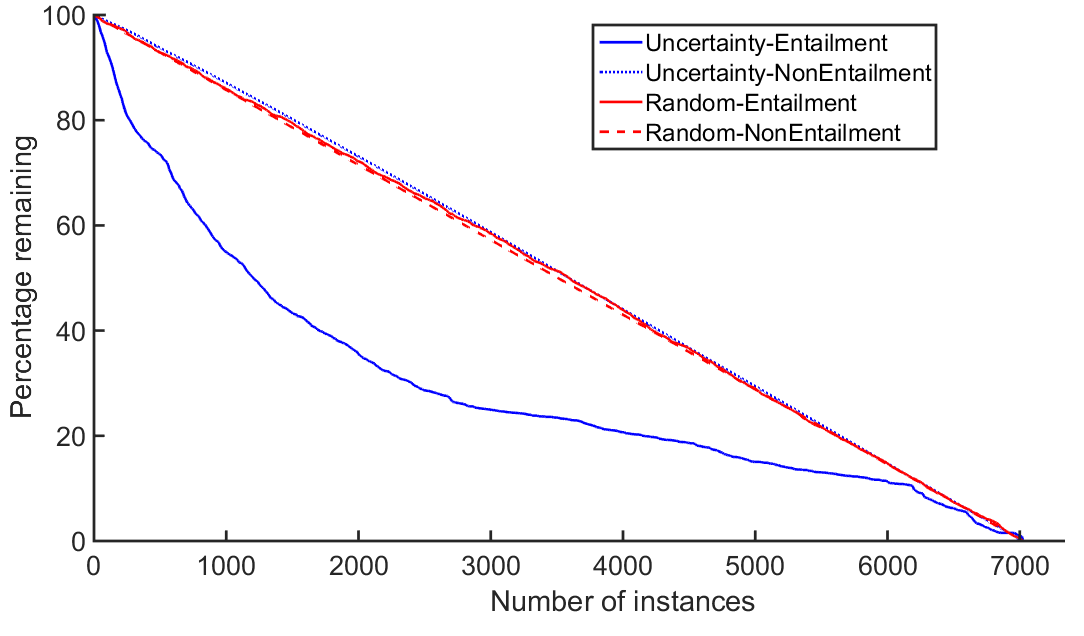}
\caption{Comparison of sampling strategies for active learning (on newswire development set)}
\label{fig:active-depletion}
\end{figure}

%\subsection{Class balance and active learning}
%\label{subsec:class_bal_active}
Finally, for active learning, we consider random sampling as a competing approach to uncertainty sampling. Figure \ref{fig:active-depletion} illustrates the percentage of positive and negative instances that get included in the training set for both sampling strategies, as active learning proceeds. The blue solid line shows that positive instances are \textit{consumed} faster  than the negative instances with uncertainty sampling. Thus, a higher percentage of positive instances (that approximately equals the number of negative instances getting added) get added and this helps maintain a balanced class distribution.

Once the positive instances are exhausted, more negative instances are added, resulting in some class imbalance that hurts performance (even though more training data is being added overall). In contrast, random sampling does not change the class balance, as it \textit{consumes} a proportional number of positive and negative instances (resulting in more negative than positive instances). The plot indicates that when using uncertainty sampling 80\% of the positive examples are added to the training set with less than 50\% of the data. This also explains how the active learner matches the performance of the model using the entire labeled set, but with fewer training examples.

%Most studies using active learning report results where the active learner matches the performance of the model using the entire labeled set, but with fewer instances of training data. We found that active learning performs better than the fully supervised model over two domains. We analyzed the possible reason for this phenomenon by tracking the proportion of positive instances to negative instances in the training set, as active learning proceeds. Figure \ref{fig:active-depletion} illustrates this on the development set of RTE data.

%The blue solid line shows that positive instances are ``consumed'' by the active learner at a faster rate than the negative instances when using uncertainty sampling. In contrast, random sampling consumes both positive and negative instances at an equal rate. Thus, the active learner tries to maintain a class balance during the initial phase of the training regime. In the later phase, more negative instances are added and thus the class imbalance hurts the performance, even though more training data is being added to the model. The plot also shows that 80\% of the positive examples are added to the training set with less than 50\% of the data.

% \begin{table}
% \centering
% \begin{tabular}{cccc}
% \toprule
% \textbf{Dataset} & \textbf{Partial} & \textbf{Complete} & \textbf{Total} \\
% \midrule
% Train		& 171	& 122	& 293	\\
% Dev		& 264	& 60		& 324	\\
% Test		& 481	& 115	& 596	\\
% \bottomrule
% \end{tabular}
% \caption{Distribution of partial and complete entailments in the corpus.}
% \label{tab:partialvscomplete}
% \end{table}

\section{Conclusion}
\label{sec:conclusion}

We explored the problem of textual entailment search in two domains -- newswire and clinical -- and focused a spotlight on the cost of obtaining labeled data in certain domains. In the process, we first created an entailment dataset for the clinical domain, and a highly competitive supervised entailment system, called {\sc Ent}, which is effective (out of the box) on two domains. We then explored two strategies -- self-training and active learning -- to address the lack of labeled data, and observed some interesting results. Our self-training system substantially improved over {\sc Ent}, achieving an F-score gain of 15\% on newswire and 13\% on clinical, using only additional unlabeled data. On the other hand, our active learning experiments demonstrated that we could match (and even beat) the baseline {\sc Ent} system with only 6.6\% of the training data in the clinical domain, and only 5.8\% of the training data in the newswire domain.
%We outline the issues while constructing a clinical dataset, for this task. Our results show that although available systems developed for textual entailment work well on newswire text, they fail to do so on clinical data.
%Our system performs well  in both domains in a supervised setting. Additionally, we show that unlabeled data can be used effectively to boost the performance of this system using semi-supervised and active learning approaches. The simple self-training algorithm works surprisingly well for the task and active learning with uncertainty sampling demonstrates that optimal performance can be achieved with less than half of the training data. These techniques leveraging unlabeled data are shown to be effective in both news and clinical domains. 

\section*{Acknowledgments}
We thank our in-house medical expert, Jennifer Liang, for guidance on the data annotation task, our medical annotators %-- ... --% 
for annotating clinical data for us, and Murthy Devarakonda for valuable insights during the project. We also thank Eric Fosler-Lussier and Albert M. Lai for their help in conceptualizing this work.% Many thanks to the anonymous reviewers for comments and suggestions on this work.

%\bibliography{acl2016}
\bibliography{acl2016}

\begin{thebibliography}{}

\bibitem[\protect\citename{Bentivogli \bgroup et al.\egroup }2009]{RTE5}
Luisa Bentivogli, Ido Dagan, Hoa~Trang Dang, Danilo Giampiccolo, and Bernardo
  Magnini.
\newblock 2009.
\newblock {The Fifth PASCAL Recognizing Textual Entailment Challenge}.
\newblock In {\em Proceedings of the Second Text Analysis Conference},
  Gaithersburg, MD.

\bibitem[\protect\citename{Bentivogli \bgroup et al.\egroup
  }2010]{Bentivogli2010}
Luisa Bentivogli, Peter Clark, Ido Dagan, and Danilo Giampiccolo.
\newblock 2010.
\newblock {The Sixth PASCAL Recognizing Textual Entailment Challenge}.
\newblock In {\em Proceedings of the Third Text Analysis Conference},
  Gaithersburg, MD.

\bibitem[\protect\citename{Bentivogli \bgroup et al.\egroup
  }2011]{Bentivogli2011}
Luisa Bentivogli, Peter Clark, Ido Dagan, and Danilo Giampiccolo.
\newblock 2011.
\newblock {The Seventh PASCAL Recognizing Textual Entailment Challenge}.
\newblock In {\em Proceedings of the Fourth Text Analysis Conference},
  Gaithersburg, MD.

\bibitem[\protect\citename{Bodenreider}2004]{Bodenreider2004}
Olivier Bodenreider.
\newblock 2004.
\newblock {The Unified Medical Language System (UMLS): Integrating Biomedical
  Terminology.}
\newblock {\em Nucleic Acids Research}, 32(Database Issue):D267--D270.

\bibitem[\protect\citename{Bowman \bgroup et al.\egroup }2015]{bowman2015}
Samuel Bowman, Gabor Angeli, Christopher Potts, and Christopher Manning.
\newblock 2015.
\newblock A large annotated corpus for learning natural language inference.
\newblock In {\em Proceedings of the 2015 Conference on Empirical Methods in
  Natural Language Processing}, pages 632--642, Lisbon, Portugal.

\bibitem[\protect\citename{Burger and Ferro}2005]{Burger2005}
John Burger and Lisa Ferro.
\newblock 2005.
\newblock {Generating an Entailment Corpus from News Headlines}.
\newblock In {\em Proceedings of the ACL Workshop on Empirical Modeling of
  Semantic Equivalence and Entailment}, pages 49--54, Ann Arbor, MI.

\bibitem[\protect\citename{Celikyilmaz \bgroup et al.\egroup
  }2009]{Celikyilmaz2009}
Asli Celikyilmaz, Marcus Thint, and Zhiheng Huang.
\newblock 2009.
\newblock {A Graph-based Semi-Supervised Learning for Question-Answering}.
\newblock In {\em Proceedings of the Joint Conference of the 47th Annual
  Meeting of the ACL and the 4th International Joint Conference on Natural
  Language Processing of the AFNLP}, pages 719--727, Singapore.

\bibitem[\protect\citename{Chawla \bgroup et al.\egroup }2002]{chawla2002smote}
Nitesh~V. Chawla, Kevin~W. Bowyer, Lawrence~O. Hall, and W.~Philip Kegelmeyer.
\newblock 2002.
\newblock {SMOTE: Synthetic Minority Over-sampling Technique}.
\newblock {\em Journal of Artificial Intelligence Research}, 16:321--357.

\bibitem[\protect\citename{Dagan \bgroup et al.\egroup }2013]{dagan2013}
Ido Dagan, Dan Roth, Mark Sammons, and Fabio~Massimo Zanzotto.
\newblock 2013.
\newblock {Recognizing Textual Entailment: Models and Applications}.
\newblock {\em Synthesis Lectures on Human Language Technologies}, 6(4):1--220.

\bibitem[\protect\citename{Hall \bgroup et al.\egroup }2009]{Hall2009}
Mark Hall, Eibe Frank, Geoffrey Holmes, Bernhard Pfahringer, Peter Reutemann,
  and Ian~H. Witten.
\newblock 2009.
\newblock {The WEKA Data Mining Software}.
\newblock {\em ACM SIGKDD Explorations Newsletter}, 11(1):10.

\bibitem[\protect\citename{Hickl and Bensley}2007]{Hickl2007}
Andrew Hickl and Jeremy Bensley.
\newblock 2007.
\newblock {A Discourse Commitment-based Framework for Recognizing Textual
  Entailment}.
\newblock In {\em Proceedings of the Workshop on Textual Entailment and
  Paraphrasing}, pages 171--176.

\bibitem[\protect\citename{Hickl \bgroup et al.\egroup }2006]{Hickl2006}
Andrew Hickl, Jeremy Bensley, John Williams, Kirk Roberts, Bryan Rink, and Ying
  Shi.
\newblock 2006.
\newblock {Recognizing Textual Entailment with LCC’s Groundhog System}.
\newblock In {\em Proceedings of the Second PASCAL Challenges Workshop}.

\bibitem[\protect\citename{Huang and Riloff}2012]{huang2012bootstrapped}
Ruihong Huang and Ellen Riloff.
\newblock 2012.
\newblock {Bootstrapped Training of Event Extraction Classifiers}.
\newblock In {\em Proceedings of the 13th Conference of the European Chapter of
  the Association for Computational Linguistics}, pages 286--295, Avignon,
  France.

\bibitem[\protect\citename{Japkowicz}2000]{japkowicz2000learning}
Nathalie Japkowicz.
\newblock 2000.
\newblock {Learning from Imbalanced Data Sets: A Comparison of Various
  Strategies}.
\newblock In {\em AAAI Workshop on Learning from Imbalanced Data Sets}, pages
  10--15.

\bibitem[\protect\citename{Jia \bgroup et al.\egroup }2010]{Jia2010}
Houping Jia, Xiaojiang Huang, Tengfei Ma, Xiaojun Wan, and Jianguo Xiao.
\newblock 2010.
\newblock {PKUTM Participation at TAC 2010 RTE and Summarization Track}.
\newblock In {\em Proceedings of the Third Text Analysis Conference},
  Gaithersburg, MD.

\bibitem[\protect\citename{Kouylekov and Negri}2010]{Kouylekov2010}
Milen Kouylekov and Matteo Negri.
\newblock 2010.
\newblock {An Open-Source Package for Recognizing Textual Entailment}.
\newblock In {\em Proceedings of the ACL 2010 System Demonstrations}, Uppsala,
  Sweden.

\bibitem[\protect\citename{Lewis and Gale}1994]{Lewis94}
David~D. Lewis and William~A. Gale.
\newblock 1994.
\newblock {A Sequential Algorithm for Training Text Classifiers}.
\newblock In {\em Proceedings of the 17th Annual International ACM SIGIR
  Conference on Research and Development in Information Retrieval}, pages
  3--12, Dublin, Ireland.

\bibitem[\protect\citename{Magnini \bgroup et al.\egroup }2014]{Magnini2014}
Bernardo Magnini, Roberto Zanoli, Ido Dagan, Kathrin Eichler, Neumann Guenter,
  Tae-Gil Noh, Sebastian Pad\'{o}, Asher Stern, and Omer Levy.
\newblock 2014.
\newblock {The Excitement Open Platform for Textual Inferences}.
\newblock In {\em Proceedings of 52nd Annual Meeting of the Association for
  Computational Linguistics: System Demonstrations}, pages 43--48, Baltimore,
  MD.

\bibitem[\protect\citename{McClosky \bgroup et al.\egroup
  }2006]{mcclosky2006effective}
David McClosky, Eugene Charniak, and Mark Johnson.
\newblock 2006.
\newblock {Effective Self-Training for Parsing}.
\newblock In {\em Proceedings of Human Language Technology Conference of the
  North American Chapter of the Association of Computational Linguistics},
  pages 152--159, New York City, NY.

\bibitem[\protect\citename{McCord}1989]{mccord89}
Michael McCord.
\newblock 1989.
\newblock {Slot Grammar: A System for Simpler Construction of Practical Natural
  Language Grammars}.
\newblock In {\em Proceedings of the International Symposium on Natural
  Language and Logic}, pages 118--145.

\bibitem[\protect\citename{McInnes \bgroup et al.\egroup }2009]{McInnes2009}
Bridget~T McInnes, Ted Pedersen, and Serguei V.~S. Pakhomov.
\newblock 2009.
\newblock {UMLS-Interface and UMLS-Similarity : Open Source Software for
  Measuring Paths and Semantic Similarity.}
\newblock In {\em Proceedings of the Annual Symposium of the American Medical
  Informatics Association}, San Francisco, CA.

\bibitem[\protect\citename{Mihalcea}2004]{mihalcea2004co}
Rada Mihalcea.
\newblock 2004.
\newblock {Co-Training and Self-Training for Word Sense Disambiguation}.
\newblock In {\em Proceedings of the Eighth Conference on Natural Language
  Learning}, pages 33--40, Boston, MA.

\bibitem[\protect\citename{Mirkin \bgroup et al.\egroup }2009]{Mirkin2009}
Shachar Mirkin, Roy Bar-Haim, Jonathan Berant, Ido Dagan, Eyal Shnarch, Asher
  Stern, and Idan Szpektor.
\newblock 2009.
\newblock {Addressing Discourse and Document Structure in the RTE Search Task}.
\newblock In {\em Proceedings of the Second Text Analysis Conference},
  Gaithersburg, MD.

\bibitem[\protect\citename{Moschitti}2004]{moschitti06}
Alessandro Moschitti.
\newblock 2004.
\newblock {A Study on Convolution Kernels for Shallow Statistic Parsing}.
\newblock In {\em Proceedings of the 42nd Annual Meeting of the Association for
  Computational Linguistics}, pages 335--342, Barcelona, Spain.

\bibitem[\protect\citename{Patwardhan and Riloff}2007]{patwardhan2007ie}
Siddharth Patwardhan and Ellen Riloff.
\newblock 2007.
\newblock {Effective Information Extraction with Semantic Affinity Patterns and
  Relevant Regions}.
\newblock In {\em Proceedings of the 2007 Joint Conference on Empirical Methods
  in Natural Language Processing and Computational Natural Language Learning},
  pages 717--727, Prague, Czech Republic.

\bibitem[\protect\citename{Pedersen \bgroup et al.\egroup }2007]{Pedersen2007}
Ted Pedersen, Serguei V.~S. Pakhomov, Siddharth Patwardhan, and Christopher~G
  Chute.
\newblock 2007.
\newblock {Measures of Semantic Similarity and Relatedness in the Biomedical
  Domain}.
\newblock {\em Journal of Biomedical Informatics}, 40(3):288--99.

\bibitem[\protect\citename{Reichart \bgroup et al.\egroup }2008]{reichart2008}
Roi Reichart, Katrin Tomanek, Udo Hahn, and Ari Rappoport.
\newblock 2008.
\newblock {Multi-Task Active Learning for Linguistic Annotations}.
\newblock In {\em Proceedings of ACL-08: HLT}, pages 861--869, Columbus, OH.

\bibitem[\protect\citename{Settles and Craven}2008]{Settles2008}
Burr Settles and Mark Craven.
\newblock 2008.
\newblock {An Analysis of Active Learning Strategies for Sequence Labeling
  Tasks}.
\newblock In {\em Proceedings of the 2008 Conference on Empirical Methods in
  Natural Language Processing}, pages 1070--1079, Honolulu, HI.

\bibitem[\protect\citename{Settles}2012]{settles2012}
Burr Settles.
\newblock 2012.
\newblock {Active Learning}.
\newblock {\em Synthesis Lectures on Artificial Intelligence and Machine
  Learning}, 6(1):1--114.

\bibitem[\protect\citename{Shivade \bgroup et al.\egroup }2015]{Shivade2015}
Chaitanya Shivade, Courtney Hebert, Marcelo Loptegui, Marie-Catherine
  de~Marneffe, Eric Fosler-Lussier, and Albert~M. Lai.
\newblock 2015.
\newblock {Textual Inference for Eligibility Criteria Resolution in Clinical
  trials}.
\newblock {\em Journal of Biomedical Informatics}, 58:S211--S218.

\bibitem[\protect\citename{Tsuchida and Ishikawa}2011]{Tsuchida2011}
Masaaki Tsuchida and Kai Ishikawa.
\newblock 2011.
\newblock {IKOMA at TAC2011 : A Method for Recognizing Textual Entailment using
  Lexical-level and Sentence Structure-level Features}.
\newblock In {\em Proceedings of the Fourth Text Analysis Conference},
  Gaithersburg, MD.

\bibitem[\protect\citename{Wang and Zhang}2009]{wang2009}
Rui Wang and Yi~Zhang.
\newblock 2009.
\newblock {Recognizing Textual Relatedness with Predicate-Argument Structures}.
\newblock In {\em Proceedings of the 2009 Conference on Empirical Methods in
  Natural Language Processing}, pages 784--792, Singapore.

\bibitem[\protect\citename{Wang \bgroup et al.\egroup }2011]{wang11}
Chang Wang, James Fan, Aditya Kalyanpur, and David Gondek.
\newblock 2011.
\newblock {Relation Extraction with Relation Topics}.
\newblock In {\em Proceedings of the 2011 Conference on Empirical Methods in
  Natural Language Processing}, pages 1426--1436, Edinburgh, UK.

\bibitem[\protect\citename{Wang \bgroup et al.\egroup }2012]{wang12}
Chang Wang, Aditya Kalyanpur, James Fan, Branimir~K. Boguraev, and David
  Gondek.
\newblock 2012.
\newblock {Relation Extraction and Scoring in DeepQA}.
\newblock {\em IBM Journal of Research and Development}, 56(3.4):9:1--9:12.

\bibitem[\protect\citename{Zanzotto and Moschitti}2006]{zanzotto2006}
Fabio~M. Zanzotto and Alessandro Moschitti.
\newblock 2006.
\newblock {Automatic Learning of Textual Entailments with Cross-Pair
  Similarities}.
\newblock In {\em Proceedings of the 21st International Conference on
  Computational Linguistics and 44th Annual Meeting of the Association for
  Computational Linguistics}, pages 401--408, Sydney, Australia.

\bibitem[\protect\citename{Zanzotto and Pennacchiotti}2010]{Zanzotto2010}
Fabio~M. Zanzotto and Marco Pennacchiotti.
\newblock 2010.
\newblock {Expanding Textual Entailment Corpora from Wikipedia using
  Co-training}.
\newblock In {\em Proceedings of the 2nd Workshop on The People's Web Meets
  NLP: Collaboratively Constructed Semantic Resources}, pages 28--36, Beijing,
  China.

\end{thebibliography}
\bibliographystyle{acl2016}

\end{document}